# Machine Translation Model based on Non-parallel Corpus and Semi-supervised Transductive Learning

**Abstract**  Although the parallel corpus has an irreplaceable role in machine translation, its scale and coverage is still beyond the actual needs. Non-parallel corpus resources on the web have an inestimable potential value in machine translation and other natural language processing tasks. This article proposes a semi-supervised transductive learning method for expanding the training corpus in statistical machine translation system by extracting parallel sentences from the non-parallel corpus. This method only requires a small amount of labeled corpus and a large unlabeled corpus to build a high-performance classifier, especially for when there is short of labeled corpus. The experimental results show that by combining the non-parallel corpus alignment and the semi-supervised transductive learning method, we can more effectively use their respective strengths to improve the performance of machine translation system.

# 1 Introduction

For most of pairs of languages, the available parallel corpus is very limited. Even with many bilingual corpora, the language pair of English and French is also confined in specific areas, such as political speech. So when training corpus in a particular field for a machine translation system, it is hard to translate the text in different fields.

In order to solve this problem, we turned to a comparable non-parallel corpus. Comparable Corpus contains two corpora in different language that are not parallel in the strict sense, but they are related, and often convey the same information. Such as Xinhua News, CNN, the BBC, which often contain sentence pairs that can be used as the translations.

Non-parallel corpus is also known as comparable corpus formed by the texts of two languages with some similarities. Such as: News belong to different sites on the same day and have the same theme both in Chinese and English. Non-parallel corpus often has texts in two languages which may be the relevant reports of the same event written by different writers using there native languages, so they are comparable. In the strict sense, non-parallel corpus is not parallel, but the messages that the two texts delivered is relevant, and the words in sentences may have different orders. For it is easy to obtain from the Internet, the non-parallel corpus has been used instead of the parallel corpus as the training corpus of machine translation. Dragos Stefan Munteanu, et al (2005) studied how to align sentences of the non-parallel corpus for a machine translation system. They first obtain the candidate translations from the non-parallel corpus by the method of information retrieval, and then using the maximum entropy classifier to accurately determine whether a sentence pair is corresponding translations. The experimental results show that the data obtained can effectively improve the accuracy of statistical machine translation.

Based on the assumption that the information provided by the non-parallel corpus can improve the accuracy of machine translation, we propose a semi-supervised transductive learning framework. "Transductive" means we repeat to translate the source language in the non-parallel corpus, then according to the translation results, we select the corresponding translations from the target language and source language sentences together to form parallel sentences in order to improve the accuracy of statistical machine translation. This approach also allows the machine translation system to adapt to texts of a new areas.

## 2 Sentence alignment of bilingual corpus

Before applying bilingual corpus to machine translation, we has to determine the corresponding paragraphs, sentences, or even phrases, single words of the source and target languages which are mutual translations. This process is known as the "alignment". In general, the unit for collecting corporal is the chapter, so the first step of processing bilingual corpora is to align sentence.

In the earlier researches of sentence alignment, we generally compare the length of sentences in two languages. The basic idea of this method is that the long sentence of the source language corresponds to the long sentences of the target language, and the short sentence of the source language corresponds to the short sentence of the target language. There are proportional relationships between the sentence length of two languages. Calculating the length of the sentence has two methods: measure the number of characters or the number of words. Both methods achieved good alignment results in the Canadian Hansards bilingual corpus of British and French. Peter F. Brown (1991) uses the word as the unit to calculate the length. He selected 90% sentences of the Canadian Hansards corpus that is easier to handle. The alignment accuracy rate achieved to 99.4%. William A. Gale (1993) used characters as the calculation unit of length. The alignment accuracy achieved to 98% on all the sentences of Canadian Hansards corpus. And for the 80% part of it which is easier to handle, the alignment accuracy achieved to 99.6%. But the alignment method based on the length is not suitable for all languages. For two languages (such as English and Chinese) in different linguistic families, the results are not satisfactory that the translations are likely to be missed (Liu Xin, 1995). Dekai Wu (1994) aligned Hong Kong Hansards Chinese-English bilingual corpus with this method. The accuracy rate is only 86.4%.

Another sentence alignment method is based on the vocabulary. Martin Kay, et al (1993) proposed that if the distribution of the two words is the same, then they are the corresponding unit. That is if a word occurs in the vast majority of a sentence, its corresponding word must also appear in the corresponding sentence. They use American Articles as the corpus. After four iterations, 96% of the sentences will be aligned. Stanley F. Chen et al (1993) adopted the EM iterative algorithm. According to the word-to-word translation model, they use the dynamic programming method to determine the optimal alignment mode. With the emergence of bilingual dictionaries, bilingual dictionaries are used to directly align sentences. Yang Muyun(2002) accomplishes the bilingual sentence alignment method based on bilingual dictionaries in English and Chinese, and the accuracy rate achieves 93%. Takehito Utsuro (1994) combined bilingual dictionaries and statistical methods to build a bilingual alignment framework. In addition, Dekai Wu (1994) proposed a hybrid-aligned technology by combining vocabulary and length. He tested his system in the Hong Kong Hansards Chinese-English bilingual corpus, and the accuracy rate achieves 92.1%.

## 3 semi-supervised machine learning methods

Machine Learning is the core of the artificial intelligence research, which usually refers to that through the induction and synthesis approach the computer simulates the human learning ability, constructs its models and algorithms to acquire new knowledge or skills from the known sample, and re-organizes existing learning structure so as to continuously improve their performance. Although the mechanism of human learning is not yet clear, by the induction of calculation model, the system can automatically make adaptive changes to complete the complex tasks which only human beings can complete originally.

Machine learning includes the supervised learning and the unsupervised learning. In traditional supervised machine learning, the learning algorithm training the model on the labeled sample set given by the outside world, inductive parameters from the sample set, and label the unknown data set based on the knowledge learned (by Trevor Hastie, 2005). Most of the classification methods are supervised machine learning. However, in many applications, labeling the sample set often requires a lot of artificial and financial resources, and the process is tedious and consistency that the quality can not be guaranteed. Because the model overfits the training data, the traditional supervised learning methods are often unable to build classifiers to meet the actual demand. Different from the supervised learning, unsupervised learning methods do not need data with labels, and the number of classes to learn may not know in advance, such as clustering and the EM algorithm.

Semi-supervised learning is a kind of area between supervised learning and supervised learning. It requires a small labeled training data set, and an unlabeled training data set. The labeling algorithm is as follows (Ian H. Witten, 2006):

(1) use of labeled data to train a classifier;
(2) the classifier is applied to unlabeled data, gives each element a category according to the class probability;
(3) adding the new annotation data to the training set;
(4) use the new training set to train a new classifier;
(5) iterate until convergence.

# 4 Related Research

Unsupervised learning method was first applied to the word alignment. Chris Callison -Burch, et al (2004) use unsupervised learning on a sentence-aligned parallel texts to acquire a word alignment model, and also use the unsupervised method to train another model with a small artificial word-aligned corpus. They combine the two models as word alignment statistical data, reduce the alignment error rate and improve the performance of the machine translation system. Alexander Fraser, et al (2006) use a semi-supervised learning methods on the basis of IBM model 4: use the discriminant step to improve the quality of word alignment in a small corpus that are manually aligned, apply the estimated step and maximize step alternatively in a large-scale training corpus. This method improves the translation accuracy rate of the phrase-based statistical machine translation system.

Co-training method is a weakly supervised learning method. Chris Callison-Burch (2002) introduces co-training to statistical machine translation. This approach requires a variety of source language aligned with the same target language sentence as a small amount of manually annotated initial data. They use the initial small-scale manually annotated data to obtain the large-scale annotated corpus by iteration. For source language sentence of each language, they use different translation models to obtain the target language candidates, and then select the best from the candidates as the final translation.

Nicola Ueffing (2007) proposes an transductive and adaptive training model for statistical machine translation. They use a phrase-based machine translation system to translate the source language sentences, translations with a confidence score above the threshold are added to the training set. "Transductive" means repeatedly translating sentence in development set and test set, and using the generated translations to gradually increase the performance of statistical machine translation system. The model selection step discards bad translations, and enhances the quality of the phrase translation table. Their experiments show that using the selection can obtain better results than adding all the translations to the system.

# 5 System framework

In this paper, we propose a semi-supervised transductive learning method based on non-parallel corpus. Instead of using monolingual corpus of the source language (Nicola Ueffing, 2007), we use the non-parallel bilingual corpus that can be obtained free from the Internet. Transductive learning is a semi-supervised learning method. When lack of parallel corpus, we can expand parallel corpus gradually by iteration in order to improve the system performance. Transductive learning translates the monolingual corpus by the current model and selects the translations with higher score. The selected translations are combined with their source language sentences to form a sentence pair which are added to the training corpus. Then the expanded corpus is used to train a new model to get better translations. After a dozen or even hundreds of iterations, the system performance will be gradually improved, and ultimately converge. In transductive learning, how to choose the translations with a higher accuracy rate is the key to success. Ueffing scores the translations by the priori probabilities of words and phrases and the target language model. This is a self-learning method. By discarding low-scoring translation, we can further improve the quality of the translation model.

(1) Input: a sentence-aligned parallel training set $L$
(2) Input: bilingual non-parallel corpus $U$
(3) Input: number of iterations $R$, the value of the $n-best$ list

(4) $T_{-1} := \{\}$ // Additional bilingual training corpus

(5) $i := 0$ // number of iterations
(6) repeat

(7) Training Steps: $\pi^{(i)} := Estimate(L, T_{i-1})$ // $\pi^{(i)}$ for the system parameters

(8) $X_i := \{\}$ // the generated translation set of this iteration

(9) for each source language sentence $s \in U$

(10) Labeling Step: use $\pi^{(i)}$ decoding s, compare with the corresponding target language text of non-parallel corpus, and get N sentences with the best scores.

(11) $X_i := X_i \cup \left\{ (t_n, s_n, \pi^{(i)}(t_n | s_n))_{n=1}^{N} \right\}$

(12) end

(13) Scoring Step: $S_i := Score(X_i)$ // give a score to each sentence in X

(14) Selecting Steps: $T_i := Select(X_i, S_i)$ // select a subset of X including good translations.
(15) $i := i+1$
(16) until $i > R$

Algorithm 1: semi-supervised transductive learning of non-parallel corpus

On the basis of Nicola Ueffing (2007), we propose a transductive learning method on non-parallel corpus. Different form Nicola Ueffing (2007), we compare the translations of the model with the target language in non-parallel corpus, and select correct ones adding to the training corpus. The selection process is no longer dependent on the language model and the priori probability. Instead it uses the translations of the current model as a scoring scale to align source language sentences and target language sentences in non-parallel corpus of in order to obtain new parallel sentences. This approach can take full advantage of the translation results of the current model. Moreover, the parallel sentences added the training corpus to come from the corpus itself, so the translations have the higher accuracy in favor of the system performance improvement. Transductive Learning on non-parallel corpus for the first time consider the sentence alignment and the translation system as a unified whole. Sentence alignment is based on translations produced by the translation model, and updating the translation model is based on results of sentence alignment. The specific algorithm is shown in Algorithm 1.

# 6 System Implementation

## 6.1 The method for candidate parallel sentence acquisition

Through the search engines of some portals, we can find large quantities of the corresponding Chinese and English news by several designed keywords. However, without the corresponding relationship of web pages in URL, when aligning chapters, the news cannot be directly regarded as a non-parallel corpus. Referring to Dragos Stefan Munteanu (2005), we firstly segment the Chinese web pages, and then convert each Chinese word into the corresponding English translation by bilingual dictionaries to generate English query documents.

We use Lemur information retrieval tools (http://www.lemurproject.org/) to retrieve the English documents, and return the first 20 English documents corresponding to a Chinese document. These results are returned as candidate non-parallel corpus for transductive learning.

## 6.2 Scoring Steps

According to the training model of the current step, we decode the source language texts in non-parallel. By comparing the acquired translations and the target language in non-parallel corpus, the similarity degrees are calculated as the score to decide whether a target language sentence is the corresponding translation. Since the training data are little at the start, the translation accuracy is not high, and the similarity of the translation and target translation is low, we considered the following factors synthetically:
(1) the ratio between the translation length and the target language sentence length
(2) the number of the same or similar words between the translation and the target language sentence
(3) the morphological changes of English words
(4) proper nouns (place or names) matched
(5) number matched
  According to the importance, all these factors are added with weights as the total score.

## 6.3 Selecting Step

By comparing with the translations of candidate parallel corpus, the selected English sentences have to meet the following two conditions to compose the sentence pairs with translations.
(1) the score must be higher than the preset threshold;
(2) the score must be the highest in all matching sentences.
The threshold setting directly impacts on the quantity and quality of translations. Moreover, since the number of matched translation sentences may change with the length of the sentence, for different lengths of English sentences, we set a different threshold.

## 6.4 non-parallel corpus example

From Xinhua news sites, we get a total of 994 Chinese-English texts. The following lists one of them.

【text】

**Chinese（25 sentences）：**
1. 亚洲 股市 受 美 鼓舞 星期二 强劲 反弹 。
2. 亚洲 证券 市场 连续 第二 天 股市 上扬 。
3. 投资者 乐观 地 认为 ， 各国 政府 的 救 市 努力 将 挽救 处于 病态 的 全球 金融 体系 。
4. 日本 股市 暴涨 百分之 14 。
5. 在 澳大利亚 ， 投资者 受到 陆 克 文 总理 用 数十亿 美元 加强 经济 的 计划 的 鼓舞 。
6. 星期二 ， 日本 股市 是 亚洲 各地 股市 中 上涨 幅度 最 大 的 ， 日经指数 暴涨 百分之 14 ， 一 共 涨 了 1100百 多 点 。
7. 上 星期五 ， 日本 股市 暴跌 将近 百分之 10 。
8. 今天 是 有史以来 日本 股市 涨幅 最 大 的 一 天 。
9. 投资者 的 乐观 情绪 在 整个 东亚 到处 可见 。
10. 在 韩国 、 菲律宾 和 印度尼西亚 ， 股市 都 窜 升 百分之 6 以上 ， 澳大利亚 和 香港 的 关键 指数 都 上升 了 百分之 3 以上 。
11. 上星期 最后 两 天 ， 澳大利亚 股市 直线 下跌 ， 政府 决定 用 70多 亿 美元 防止 经济 衰退 ， 导致 股市 回升 。
12. 这些 资金 将来 自 预算 盈余 ， 这些 盈余 主要 是 靠 采矿 业 的 繁荣 积累 起来 的 。
13. 澳大利亚 把 大量 矿物 出口 到 中国 和 印度 。
14. 一些 经济学家 说 ， 刺激 经济 计划 可能 不足以 防止 澳大利亚 滑入 衰退 。
15. 在 这 之前 ， 澳大利亚 经历 了 17 年 的 连续 增长 。

16. 澳大利亚 总理 陆克文 发表 全国 讲话 , 他 对 未来 表示 乐 观 , 但是 他 说 , 必须 立即 采取 行动 。
17. " 全球 金融 危机 进入 了 一个 新 的 、 危险 的 、 破坏性 的 阶 段 。
18. 危机 已经 波及 实际 的 经济 领域 , 增长 和 就业 都 受到 影 响 。
19. 这 就是 为什么 政府 决定 及早 果断 行动 , 采取 针对 未来 的 经 济 安全 战略 。
20. 它 将 帮助 我们 巩固 未来 的 经济 增长 , 为 千家万户 的 未来 提供 实际 的 支持 。
21. 亚洲 股市 回升 之前 , 美国 股市 道 琼 斯 指数 星期一 上升 了 百分之 11 以上 。
22. 这 是 1933 年 以来 升幅 最 大 的 一 天 。
23. 投资者 对 美国 政府 向 银行 体系 注资 作出 反应 。
24. 注资 是 为了 刺激 贷款 。
25. 由于 全球 信贷 冻结 , 造成 现金 来源 枯竭 , 银行 无法 贷款 。

**English（15 sentences）：**
1. Asian stocks rally as optimism grows over global rescue plans
2. Asian stock markets have risen for a second consecutive day on optimism that government rescue efforts will heal the stricken global financial system .
3. Japanese stocks soared by 14 percent .
4. In Australia , investors were encouraged by prime minister Kevin Rudd ' s plan to spend billions of dollars to strengthen the economy .
5. Japan lead the gains in Asian markets tuesday as the benchmark Nikkei 225 index surged by more than 1,100 points , or 14 percent - a stunning reversal after plunging nearly 10 percent friday .
6. It was the Nikkei ' s biggest single-day gain in history .
7. The positive mood has been repeated across the region .
8. Stock prices in South Korea , the Philippines and Indonesia jumped by more than six percent , and key indexes in Australia and Hong Kong rallied more than three percent .
9. The recovery in Australian stocks , which went into freefall at the end of last week , followed a decision by the government to spend more than $ 7 billion to try to stave off recession .
10. The funds will come from the budget surplus , which has been amassed largely on the back of a mining boom that has seen vast quantities of australian minerals exported to China and India .
11. Some economists say the stimulus package may not be enough to prevent Australia slipping into a recession after 17 years of uninterrupted growth .
12. In a nationwide address , Australian Prime Minister Kevin Rudd was optimistic about the future , but said that immediate action was needed .
13. " The global financial crisis has entered into a new , dangerous and damaging phase , one which goes to the real economy , growth and jobs , " said Mr. Rudd .

" And that is why the government has decided to act decisively and early on the question of this economic security strategy for the future ; an economic security strategy to help underpin positive economic growth into the future and to provide practical support for households . "
14. The recovery of Asian stock markets followed a surge in the dow jones industrial average , which gained more than 11 percent monday - its biggest one-day gain since 1933 .
15. Investors were reacting to efforts by the U.S. government to inject capital into banking system to stimulate lending , which has dried up in the global credit meltdown .

Table 1 the relations between English sentences and Chinese sentences in the text

| English Sentence No. | 1 | 2 | 3 | 4 | 5 | 6 | 7 | 8 | 9 | 10 | 11 | 12 | 13 | 14 | 15 |
|---|---|---|---|---|---|---|---|---|---|---|---|---|---|---|---|
| Chinese Sentence No. | 1 | 2 3 | 4 | 5 | 6 7 | 8 | 9 | 10 | 11 | 12 13 | 14 15 | 16 | 17 18 19 20 | 21 22 | 23 24 25 |

The above report is a non-parallel text. The contents of English and Chinese versions describe a same news event. However the English and Chinese sentences are not strictly one-to-one correspondent, and sometimes an English sentence corresponds to a number of Chinese sentences. The corresponding relations are shown in Table 1. In the report, The Chinese part has a total of 25 sentences, while the English part has 15. Based on text analyzing, there are two reasons for different numbers of sentencese between English part and Chinese part: (1) the contents in Chinese and English are not entirely consistent (2) there are some long sentences in the English part. However, in the above texts, there are a number of corresponding pairs of parallel sentences, such as:

（1） 投资者 的 乐观 情绪 在 整个 东亚 到处 可见 。

The positive mood has been repeated across the region.

（2） 日本 股市 暴涨 百分之 14 。

Japanese stocks soared by 14 percent .

In these sentence pairs, some are corresponding in the strict sense, such as (2); but some are only roughly close in meaning, such as (1). However, whether or not they are the parallel sentence in the strict sense, as long as with the high similarity, they can be used as a useful supplement for machine translation system training corpus. In the sentences, there are some corresponding phrases, which can be added to the phrase translation table of the machine translation system.

(1):

| 乐观 | <——> | positive |
| --- | --- | --- |
| 情绪 | <——> | mood |
| 到处可见 | <——> | has been repeated across |

(2):

| 日本 | <——> | Japanese |
| --- | --- | --- |
| 股市 | <——> | stocks |
| 暴涨 | <——> | soared |
| 百分之 | <——> | percent |

## 6.5 The Acquisition of parallel sentence pairs

In semi-supervised transductive system on non-parallel corpus, we have to filter out the truly parallel sentences from the candidate parallel sentence pairs. For example, the sentence "美国 总统 布什 说 ， 美国 正在 跟 其它 国 家 合作 恢复 国际 金融 市场 的 力量 和 稳定 。" is translated by the initial translation system as:

The president bush (former us president) said, of the american wrestling with the other countries to work together to international financial market-place of the power and stabilization.

The translations "president bush" "other countries" "international financial" "and" appear in the corresponding English sentence. Moreover, the translation sentence and the English sentence are all included in the corresponding text of the news. Therefore, we can determine that they are likely to be corresponding sentence pair.

We compare the translation with the English sentence in non-parallel corpus, and figure out the number of the same n-gram words. The higher the number, the higher the coefficient is. Then the result is divided by the length factor, to get the sentence similarity score.

$$句子相似度 = \frac{\alpha_1 C_1 + \alpha_2 C_2 + \cdots + \alpha_n C_n + \beta_1 C_{数字} + \beta_2 C_{专有名词}}{\theta_{length}} \quad (2.1)$$

Among them, $\alpha_1$、$\alpha_2 \ldots \alpha_n$ are adjustment parameters for 1-gram, 2-gram ... n-gram language model which can be adjusted according to actual situation ($\alpha_1 < \alpha_2 < \cdots < \alpha_n$). $\beta_1$ And $\beta_2$ are the adjustment parameters of the numbers and proper nouns. $\theta_{length}$ Is the length factor.

# 7 Experimental Results

Because there are less non-parallel corpus that can be obtained from the Internet, so we only carry out the experiment of small-scale transductive learning. The initial corpus of the experiments is 1000 sentences selected from the aligned bilingual corpus parallel produced by Institute of Computing Technology, Chinese Academy of Sciences. Non-parallel corpus contains 994 Chinese-English news obtained from the Xinhua net, including 12,326 Chinese sentences and 283,058 English sentences. In addition, the parallel corpus contains 187,014 translation dictionaries. Language model uses a large-scale English monolingual corpus, which contains 113,723 English sentences and 1,301,274 English words.

Table 2 The evaluation results of transductive learning on non-parallel corpus-based

| Iteration | parallel corpus size | NIST2008 | | CWMT2008 | |
| --- | --- | --- | --- | --- | --- |
| | | NIST | BLEU | NIST | BLEU |
| 0 (baseline) | 1000 | **4.1758** | 6.63 | 4.3050 | 6.24 |
| 1 | 1175 | 4.1439 | 6.67 | 4.3111 | 5.91 |
| 2 | 1303 | 4.0772 | 6.93 | **4.3542** | 6.39 |
| 3 | 1388 | 4.0836 | **7.18** | 4.3117 | 6.26 |
| 4 | 1455 | 3.9948 | 7.05 | 4.0275 | 6.21 |
| 5 | 1588 | 4.0683 | **7.11** | **4.3355** | 6.41 |
| 6 | 1545 | 4.0175 | 7.07 | 4.2388 | **6.58** |
| 7 | 1573 | 4.0228 | 7.05 | 4.3093 | 6.46 |
| 8 | 1622 | 4.0245 | 6.93 | 4.2611 | 6.47 |
| 9 | **1763** | 3.9580 | 6.93 | 4.3099 | 6.39 |
| 10 | **1653** | 4.0693 | 7.01 | **4.3359** | **6.54** |

The parameter training corpus is Chinese-English translation corpus provided by NIST2008 evaluation, including a total of 691 Chinese sentences. Each Chinese sentence corresponds to four English translations. for the evaluation of system performance, we not only use NIST2008 test set, but also the English test corpus produced by the 4[th] machine translation seminar (CWMT2008), in which there are 1006 Chinese sentences.

Table 2 shows the experimental results of the semi-supervised transductive learning on non-parallel corpus after 10 iterations (including the initial system). After 10 iterations, the size of the parallel corpus gradually expand, and the number of the sentence increases from the 1000 to 1653. Among them, 653 parallel sentence pairs are obtained by the transductive learning. The BLEU scores of the two test sets are improved significantly. On NIST2008 test set, the BLEU score is 0.38, with an increase of 5.73%. The NIST score and BLEU score of CWMT2008 are also significantly increased by 0.0309 and 0.30 respectively, and the increase rate is 0.72% and 4.81% respectively. However, the best result obtained by the experiment is not in the 10[th] round, but in the 2[nd], 3[rd] and 6[th] rounds. The reason for this phenomenon is due to the parallel corpus obtained by the transductive learning may not be necessarily beneficial for improving the performance of the system, which also contains some non-corresponding sentences. For example in section 8, the sentence "美国 与 洪都拉斯 多年 来 关系 密切 。" is translated as "U.S. embassy in Honduras many years in. the close relations." The System mismatches it with the English sentence "U.S. embassy in Honduras calls for peace through dialogue", for the original translation and the English sentence have 4 fully matching words (including proper nouns), and the length of the sentence is short.

Synthesizing all the experimental results, except for the NIST score on NIST2008, the BLEU score on NIST2008 and the BLEU score and the NIST score on CWMT2008 has improved significantly. The experimental results show that in the case of small training corpus the semi-supervised transductive learning can extract parallel sentence pairs from the non-parallel corpus, gradually increase the size of the training data, so as to improve the performance of the machine translation systems.

# 8 the significance and limitations for non-parallel corpus used in machine translation

The use of non-parallel corpus opens up a new idea for training data acquisition in statistical machine translation. The semi-supervised transductive learning methods on non-parallel can gradually increase the amount of parallel sentences for machine translation systems in the absence of a ready-made training corpus. Different from adding sentence pairs with higher score to the training corpus by Nicola Ueffing et al (2007), the translations with high similarities are better for improving performance of the machine translation system. Some parallel sentence pairs contains many valuable corresponding translations, which is same as the parallel corpus requiring a lot of manpower and resources.

However, there are still some difficulties and shortcomings in practical applications for non-parallel corpus:

(1) In non-parallel corpus, real parallel sentence pairs are relatively small, increasing the difficulty to automatically obtain from the system translations. From 994 non-parallel corpora, we can only extract 763 pairs of parallel sentences, in which some are not in strict sense of the corresponding. Therefore, enabling the machine translation system performance improved significantly requires a lot of non-parallel corpus.

(2) When extracting the corresponding parallel sentence pairs from the non-parallel corpus, due to the defects of the initial translation, the system will map some sentence pairs which are in some sense similar, but not parallel sentence pairs. This makes a lot of noise in the training corpus, is bad for us to improve the translation system performance.

(3) The non-parallel corpus online which can be take advantage of is limited to news. Other areas of non-parallel corpus is almost invisible. So the use of non-parallel corpus has strict field restrictions.

Even so, for constructing the parallel corpus is time-consuming and laborious, the prospect of non-parallel corpus is still very impressive. Extracting parallel sentence pairs from non-parallel corpus can be used for machine translation systems resources acquisition as a new direction. At present, the demand for bilingual corpus is the growing, and the use of the bilingual corpus has spread to word segmentation, POS tagging, parsing, information retrieval and text classification. Therefore, we should make best use of the resources available online to obtain more parallel corpus. Non-parallel corpus is not readily available bilingual corpus. Researchers can use the computer automatic or semi-automatic method to turn the non-parallel corpus into a bilingual corpus. We believe that this approach is helpful to training data acquisition for machine translation, and can reduce the workload and time for constructing the parallel corpus.